\documentclass[conference]{IEEEtran}
\IEEEoverridecommandlockouts
\usepackage{cite}
\usepackage{amsmath,amssymb,amsfonts}
\usepackage{algorithmic}
\usepackage{graphicx}
\usepackage{textcomp}
\usepackage{xcolor}
\usepackage{caption, subfig}
\usepackage{amsmath}
\usepackage{amssymb}
\usepackage{cases}
\usepackage{multirow}
\usepackage{multicol}
\usepackage{adjustbox}
\usepackage{makecell}
\usepackage{diagbox}

\def\BibTeX{{\rm B\kern-.05em{\sc i\kern-.025em b}\kern-.08em
    T\kern-.1667em\lower.7ex\hbox{E}\kern-.125emX}}

\begin{document}


\title{Finding the Missing Data: A BERT-inspired Approach Against Package Loss in Wireless Sensing}

\author{%
    \IEEEauthorblockN{Zijian Zhao\IEEEauthorrefmark{1}\IEEEauthorrefmark{2}, Tingwei Chen\IEEEauthorrefmark{1}, Fanyi Meng\IEEEauthorrefmark{1}\IEEEauthorrefmark{3}, Hang Li\IEEEauthorrefmark{1}, Xiaoyang Li\IEEEauthorrefmark{1}, Guangxu Zhu\IEEEauthorrefmark{1} \thanks{Corresponding author: Guangxu Zhu}}%
    \IEEEauthorblockA{\IEEEauthorrefmark{1} Shenzhen Research Institute of Big Data \\ %
    \IEEEauthorrefmark{2} School of Computer Science and Engineering, Sun Yat-sen University \\ %
    \IEEEauthorrefmark{3} School of Science and Engineering, The Chinese University of Hong Kong, Shenzhen\\ %
    Email:\{zhaozj28\}@mail2.sysu.edu.cn,\{tingweichen,fanyimeng\}@link.cuhk.edu.cn,\{hangdavidli,lixiaoyang,gxzhu\}@sribd.cn
    }%
   
}

\maketitle

\begin{abstract}

Despite the development of various deep learning methods for Wi-Fi sensing, package loss often results in noncontinuous estimation of the Channel State Information (CSI), which negatively impacts the performance of the learning models. To overcome this challenge, we propose a deep learning model based on Bidirectional Encoder Representations from Transformers (BERT) for CSI recovery, named CSI-BERT. CSI-BERT can be trained in an self-supervised manner on the target dataset without the need for additional data. Furthermore, unlike traditional interpolation methods that focus on one subcarrier at a time, CSI-BERT captures the sequential relationships across different subcarriers. Experimental results demonstrate that CSI-BERT achieves lower error rates and faster speed compared to traditional interpolation methods, even when facing with high loss rates. Moreover, by harnessing the recovered CSI obtained from CSI-BERT, other deep learning models like Residual Network and Recurrent Neural Network can achieve an average increase in accuracy of approximately 15\% in Wi-Fi sensing tasks. The collected dataset WiGesture and code for our model are publicly available at https://github.com/RS2002/CSI-BERT.
\end{abstract}


\section{Introduction}

In recent years, wireless sensing is expected to be applied in ubiquitous signal-based scenarios \cite{zhu2023pushing}, including gesture recognition \cite{abdelnasser2015wigest}, human motion identification \cite{wen2023task, liu2022toward}, and target location \cite{li2023integrated}. Note that, a key issue confronting Wi-Fi sensing applications is the uneven temporal distribution of Wi-Fi Channel State Information (CSI) signals, even when a fixed sampling rate is set.

This issue is generally caused by packet loss. In practice, the receiver fails to successfully decode the packets because of factors such as weak signal strength, frequency interference, and hardware errors. Consequently, the number of data packets collected often falls below expectations, and the packet loss situation varies from second to second. This leads to the uneven data of time series. Since machine learning methods typically require input signals to possess consistent dimensions, this inconsistency presents a significant trouble in model training. Unfortunately, previous researchers have rarely focused on the task of recovering lost packets, where most studies simply employ interpolation methods to maintain consistent data dimensions. Nevertheless, these methods are not specifically tailored for CSI and fail to consider the inherent relationship among transmitters, receivers, and subcarriers, potentially leading to significant discrepancies between the recovered data and the actual signal. To address this issue, we propose the use of a Deep Neural Network (DNN) as it excels at capturing information and relationships within signal data. In this paper, we aim to design a deep learning model that specifically leverages the unique characteristics of CSI to recover lost packets.

Recently, Language Models (LMs) have demonstrated excellent performance in various fields, including image recognition, music analysis, and biology research. These models are predominantly based on the Transformer architecture \cite{Transformer}. Transformers utilize a bi-directional encoder to capture global information and an auto-regressive decoder to generate coherent data. Among these models, Bidirectional Encoder Representations from Transformers (BERT) \cite{BERT} has shown exceptional performance in understanding tasks. BERT's bi-directional encoder structure and effective pre-training methods contribute to its success. Since wireless information sequences also exhibit contextual relationships like neural networks, BERT can also help extract their inner features. Previous researchers \cite{BERT-sen, BERT-loc1, BERT-loc2} have explored the application of BERT in wireless sensing tasks. However, most of these studies directly convert continuous wireless signals into discrete tokens, resulting in significant information loss. 



\begin{figure}[htp]
\centering 
\includegraphics[width=0.45\textwidth]{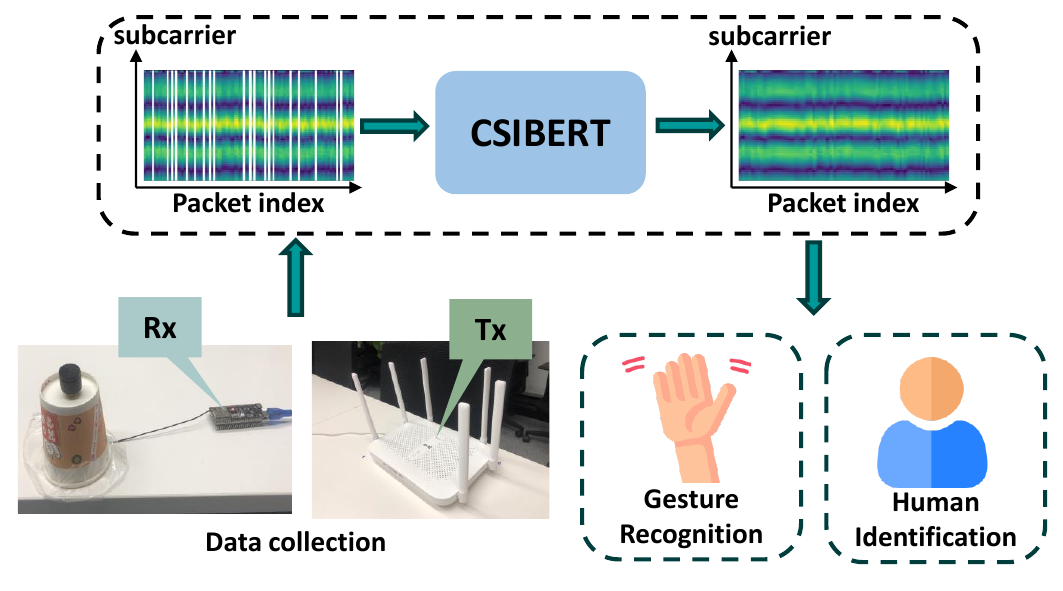}
\caption{Workflow}
\label{Workflow}
\end{figure}

In this paper, we propose a CSI-BERT model, which leverages the powerful comprehension capabilities of BERT to recover the lost CSI data. In contrast to previous BERT-based methods, we introduce a novel Embedding Layer that directly processes continuous CSI along with its corresponding time information. This approach eliminates the need to convert wireless information into discrete tokens. For practical implementation, we further present a corresponding pre-training method for CSI data, which can be trained in an self-supervised manner using target incomplete CSI sequences. To enhance the realism of the recovered CSI, we also employ adversarial learning techniques. As shown in Fig. \ref{Workflow}, our experimental system is based on the ESP32S3, a lightweight Wi-Fi sensing device, which can establish connections with home routers and extract the CSI data. The collected dataset includes identity IDs and gestures. We conduct tests on this system under various scenarios where packet loss is deliberately induced. The experimental results demonstrate that CSI-BERT exhibits outstanding performance in both CSI recovery and sensing tasks.

In summary, the main contributions of this work are:

(1) We propose CSI-BERT, the first deep learning model for CSI recovery. Our framework also provides potential directions in several domains, including the application of Natural Language Process (NLP) models in CSI, exploring pre-training methods for neural networks, and recovering multi-dimensional data.

(2) CSI-BERT can be trained exclusively on incomplete CSI data, effectively handling various packet loss rates. This capability holds significant practical implications.

(3) Experimental results demonstrate that CSI-BERT achieves lower recovery errors and faster recovery time compared to traditional interpolation methods. Moreover, it provides substantial improvements to other deep learning models like Residual Network (ResNet) \cite{Resnet} and Recurrent Neural Network (RNN) \cite{RNN} in Wi-Fi sensing tasks. Additionally, CSI-BERT itself exhibits excellent performance in CSI sensing tasks.

\section{Related Work}
BERT \cite{BERT} was initially proposed in the field of NLP. Due to its bi-directional encoder structure and pre-trained knowledge, BERT has demonstrated excellent performance in various understanding tasks. BERT employs two pre-training tasks, namely Masked Language Model (MLM) and Next Sentence Prediction (NSP). Liu et al. \cite{RoBERTa} have shown that the model achieves better performance by removing the NSP task and utilizing dynamic masking in the MLM task. Specifically, they randomly replace 15\% of the tokens with the [MASK] token during pre-training and train the model to predict the original tokens. In this work, we adopt a similar approach.

In recent years, researchers have begun applying BERT in the field of wireless sensing \cite{BERT-sen} and communication \cite{BERT-com}. However, unlike domains such as NLP and Computer Vision (CV), the field of wireless sensing lacks a sufficient amount of large datasets. Furthermore, the data in wireless sensing is heavily influenced by the environment and the collecting device, making it challenging to combine data from different datasets. Several studies have demonstrated that pretraining can enhance performance in downstream tasks, particularly in scenarios with limited training data \cite{pretrain1,pretrain2}. Consequently, pre-trained models such as BERT have been introduced to wireless sensing.

Previous researchers \cite{BERT-loc1,BERT-loc2} have predominantly focused on localization using BERT. Wang et al. \cite{BERT-map} proposed a BERT-based method for constructing radio maps. In these work, they all transformed continuous Received Signal Strength Indicator (RSSI) measurements into discrete tokens to input the data into BERT, potentially resulting in the loss of valuable RSSI information. Additionally, their approach simply adapted BERT from NLP without making significant modifications. In contrast, this paper aims to modify the BERT structure to align with the characteristics of CSI. Our proposed CSI-BERT model can directly process continuous CSI data without any loss of information.

\section{Methodology}

\begin{figure}
\centering 
\includegraphics[width=0.45\textwidth]{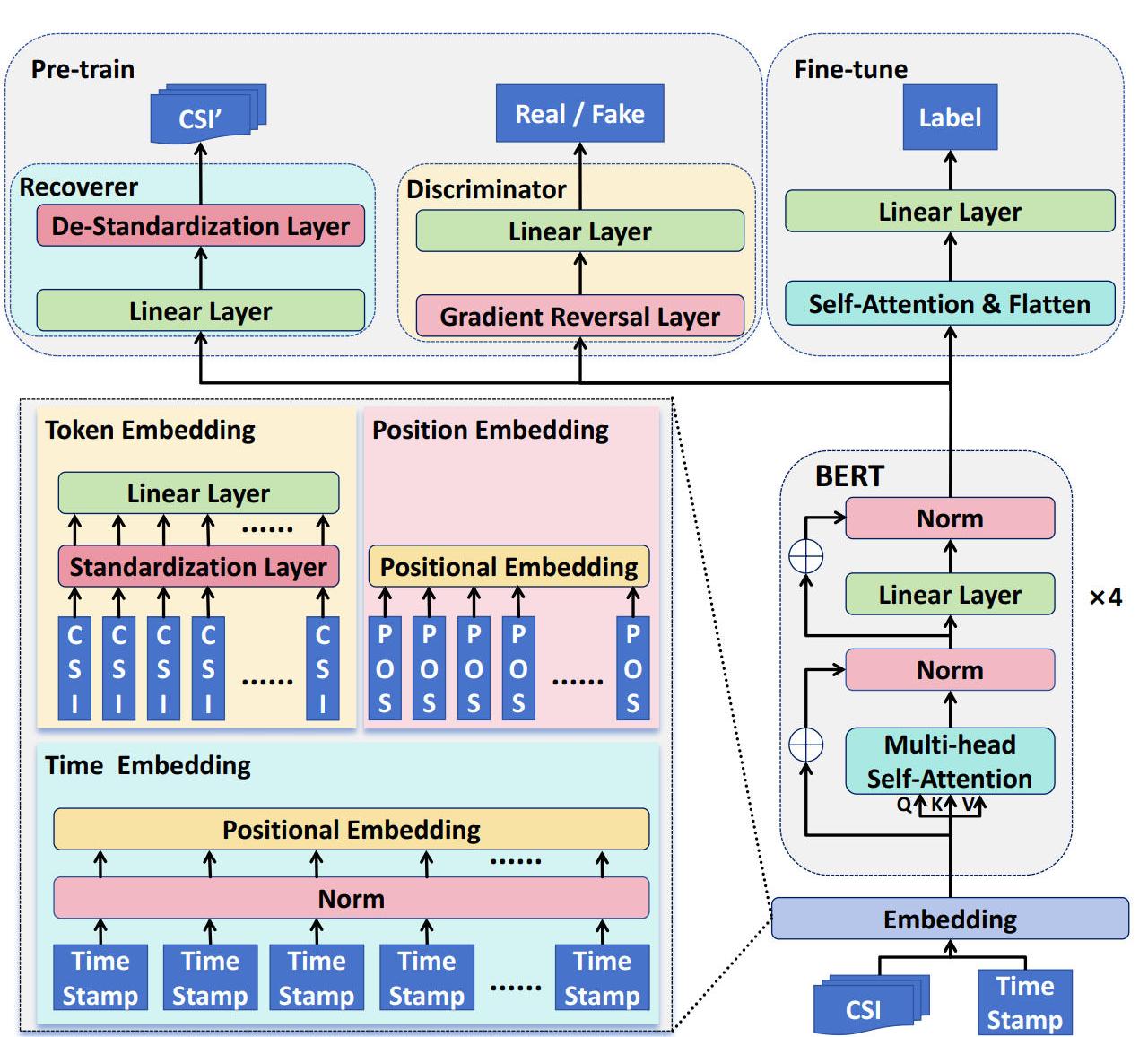}
\caption{Overview of the Proposed CSI-BERT Framework}
\label{CSI-BERT}
\end{figure}

\subsection{Overview}
The proposed CSI-BERT model is a BERT-based approach designed for CSI recovery and sensing, as illustrated in Fig. \ref{CSI-BERT}. In this approach, we leverage the CSI sequence $C=[c_1,c_2,...,c_n]$ and the corresponding timestamps $T=[t_1,t_2,...,t_n]$ as input. Each $c_i$ represents a flattened vector containing the amplitude and phase information of the CSI matrix at time $t_i$. Notably, the unit of $T$ is not a concern as it will be normalized by our model.

The utilization of CSI-BERT involves three main phases: pre-training, recovering, and fine-tuning. During pre-training, the CSI sequence is randomly destructed, and the model is trained to recover the missing components in an self-supervised manner. Subsequently, the recovery phase aims to restore the lost CSI information using two distinct strategies, namely ``recover'' and ``replace''. Finally, in the fine-tuning phase, specific tasks are employed to train the top layers of CSI-BERT for CSI Sensing tasks.

In the following sections, we provide a detailed exposition of the CSI-BERT architecture in Section \ref{Model Structure}, followed by a comprehensive description of the training and utilization steps in Section \ref{Training Step}.

\subsection{Model Structure} \label{Model Structure}
Fig. \ref{CSI-BERT} illustrates the detailed structure of CSI-BERT. CSI-BERT is built upon the BERT \cite{BERT} model, which is a bi-directional encoder based on the Transformer \cite{Transformer}. To improve the network's adaptation to the structure of CSI, we have designed novel bottom and top layers.

In the traditional BERT model, the embedding layer consists of token embedding, position embedding, and segment embedding. For the token embedding, considering that CSI is continuous, we replace the word embedding layer with a linear layer. Before the linear layer, we add a standardization layer as shown in Eq (\ref{standard}), which directly affects the effectiveness of the model.
\begin{equation}
\begin{aligned}
& \mu^{(j)}=\frac{\sum_{i=1}^n c_i^{(j)}}{n} ,\\
& \sigma^{(j)}=\sqrt{\frac{\sum_{i=1}^n (c_i^{(j)}-\mu^{(j)})^2}{n}} ,\\
& \text{Standard}(c_i^{(j)})=\frac{c_i^{(j)}-\mu^{(j)}}{\sigma^{(j)}} ,
\label{standard}
\end{aligned}
\end{equation}
where $c_i^{(j)}$ represents the $j^{th}$ dimension of $c_i$, $\mu$ and $\sigma$ represent the mean and standard deviation, respectively. These values are important for the subsequent layers.

Since CSI does not have a similar concept of a phrase or sentence, we use a time embedding to replace the segment embedding, as shown in Eq (\ref{time embedding}). To account for the varying time intervals between consecutive CSI samples, we use the time embedding to represent the absolute position, while the traditional position embedding can represent the relative position.
\begin{equation}
\begin{aligned}
& \text{TE}(t_i)^{(j)}=
    \begin{matrix}
        \begin{cases}

\text{sin}(\frac{\text{Norm}(t_i)}{10^\frac{4j}{d}}), \ j=2k\\
 \text{cos}(\frac{\text{Norm}(t_i)}{10^\frac{4(j-1)}{d}}), \ j=2k+1
        \end{cases}
    \end{matrix}
, \\
& \text{Norm}(t_i)=\frac{t_i-\text{min}(T)}{\text{max}(T)-\text{min}(T)} , 
\label{time embedding}
\end{aligned}
\end{equation}
where $j$ represents the $j^{th}$ dimension of $TE(t_i)$, $k$ is a positive integer, and $d$ is the dimension of each CSI vector $c_i$.

During pre-training and fine-tuning, we use different top layers for CSI-BERT. In the pre-training phase, the top layer consists of a Recoverer, which is used to recover the destructed CSI and is also employed during the recovering phase, and a Discriminator, which discriminates whether the CSI is real or generated by CSI-BERT.

For the Recoverer, we use a de-standardization layer to ensure that the output has a similar distribution as the input. This is achieved through the following equation:
\begin{equation}
\begin{aligned}
    \text{De-Standard}(y_i^{(j)})=(y_i^{(j)}+\mu^{(j)})*\sigma^{(j)} , 
\label{de-standard}
\end{aligned}
\end{equation}
where $y$ represents the output of the linear layer in the Recoverer, $\mu$ and $\sigma$ are calculated using Eq (\ref{standard}).

As for the Discriminator, we employ a Gradient Reversal Layer \cite{DANN} (GRL) before the final linear layer. The GRL reverses the gradient during back-propagation. The Discriminator aims to distinguish between real and generated CSI samples, but the bottom layers prevent it from classifying correctly. Through this adversarial approach, the generated CSI samples can be more realistic.

Finally, during fine-tuning, we incorporate a self-attention layer after BERT to combine the features in the time dimension. Subsequently, a linear layer is used to make the final decision.

\subsection{Key Phases} \label{Training Step}

\begin{figure*}
\centering 
\subfloat[Pre-training Phase]{\label{Pre-training Phase}\includegraphics[width=0.45\textwidth]{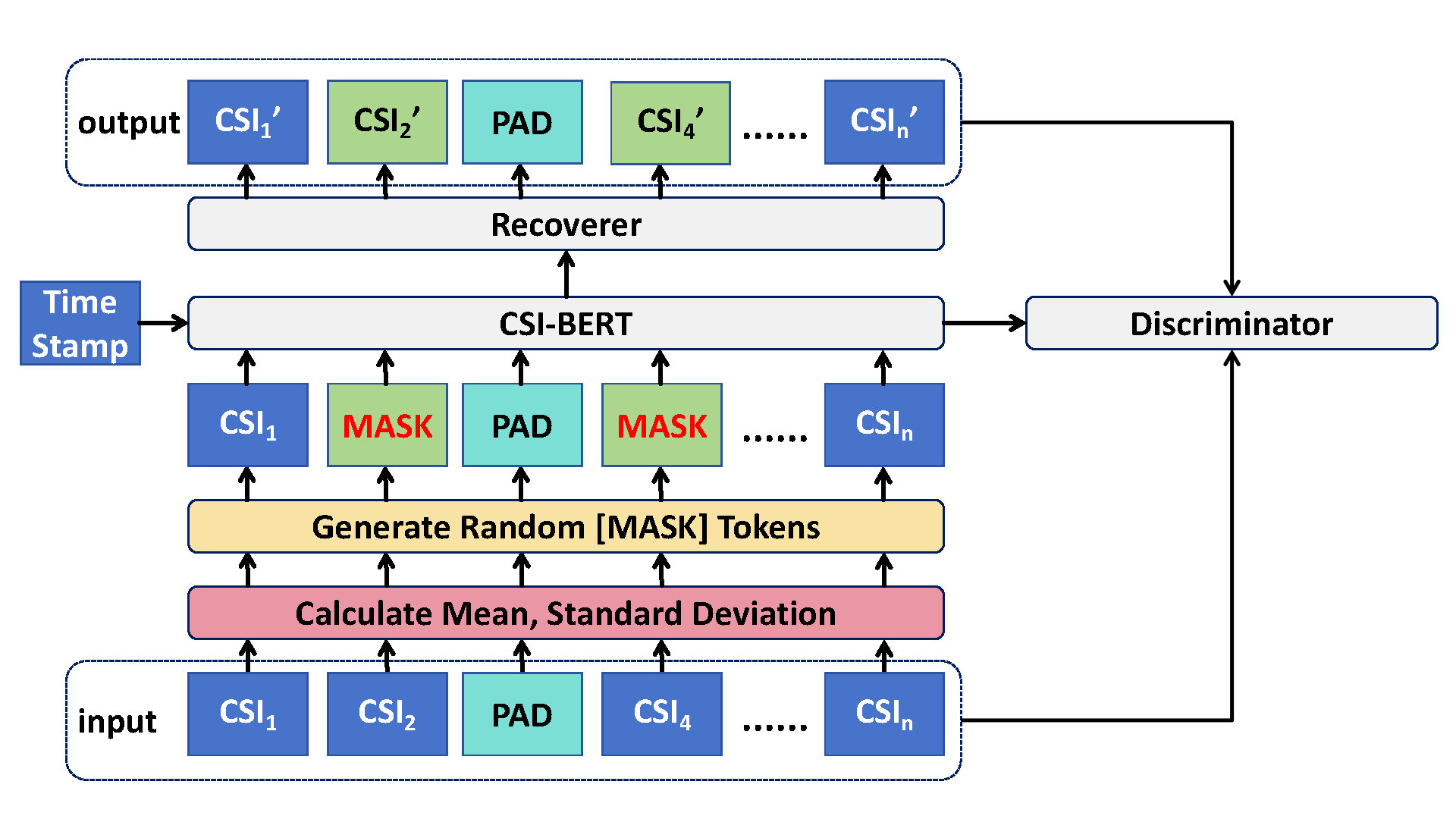}}
\subfloat[Recovering Phase]{\label{Recovery Phase}\includegraphics[width=0.45\textwidth]{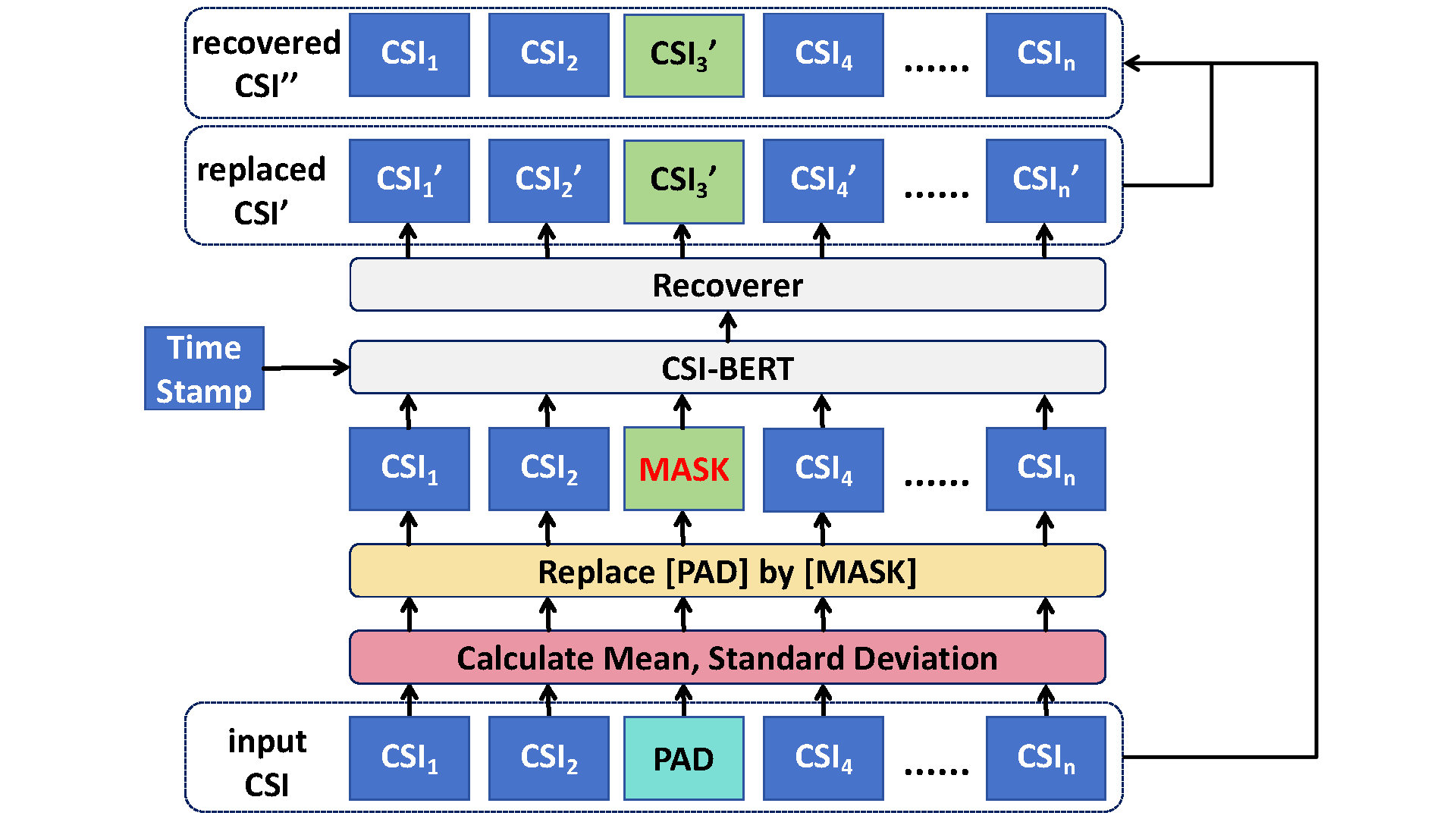}}
\caption{Process of Pre-training and Recovering}
\label{CSI-BERT-train}
\end{figure*}

\subsubsection{Pre-training Phase}
In the pre-training phase of CSI-BERT, we make use of a similar approach to the MLM task in BERT \cite{BERT}, which  involves fulfilling the empty positions by [PAD] tokens, randomly masking tokens by [MASK] tokens, and training the model to recover them. However, due to the differences between continuous CSI and discrete natural language, we have made some modifications to adapt to the CSI data.

In BERT, the [MASK] token is a fixed token, but in CSI-BERT, we replace the word embedding layer with a linear layer, which means the [MASK] token can have different values affecting the output. To address this, we assign a random value to the [MASK] token, sampled from a Gaussian distribution as shown in Eq (\ref{MASK}) whose mean and standard deviation are calculated from the standardization layer. This random approach makes it more challenging for the model to identify the position of the [MASK] token, thereby improving its understanding of the CSI sequence structure. The [PAD] token, on the other hand, can be assigned a fixed value since it is disregarded by the model with the assistance of the Attention Mask Matrix.

\begin{equation}
\begin{aligned}
[MASK]_{i}^{(j)} \sim \text{N}(\mu^{(j)},\sigma^{(j)}) , 
\label{MASK}
\end{aligned}
\end{equation}

For the pre-training phase, we follow a similar setting as RoBERTa \cite{RoBERTa} as shown in Fig. \ref{Pre-training Phase}, where the [PAD] token is used to fill the positions of missing CSI, and some non-[PAD] tokens are randomly replaced with the [MASK] token. In RoBERTa, the masking proportion is fixed at 15\%. However, in practical scenarios, the loss rate of CSI can vary, so we introduce a random masking proportion ranging from 15\% to 70\%, which is varied in each epoch.

Next, we train the model to recover these [MASK] tokens and make the recovered CSI more realistic using the Discriminator. We design a mixed loss function consisting of five parts based on the features of the CSI, as shown in Eq (\ref{Loss}).
\begin{equation}
\begin{aligned}
& L_1=\text{MSE}(C,\hat{C}) , \\
& L_2=\text{MSE}(\mu,\hat{\mu}) , \\
& L_3=\text{MSE}(\sigma,\hat{\sigma}) ,\\
& L_4=\text{CrossEntropy}(\text{Discriminator}(C),0) , \\
& L_5=\text{CrossEntropy}(\text{Discriminator}(\hat{C}),1) , \\
\label{Loss}
\end{aligned}
\end{equation}
where $\hat{C}$ represents the output of the Recoverer, $\hat{\mu},\hat{\sigma}$ represents the mean and standard deviation of $\hat{C}$ in the dimension of time, respectively. The $L_1$ loss is the traditional loss function used in BERT \cite{BERT}, as it measures the accuracy of the output. However, in CSI recovery, the overall shape of the CSI is also important. Therefore, we utilize the $L_2,L_3$ losses to consider the overall shape of the CSI by assessing its mean, and standard deviation. Furthermore, we train the Recoverer to classify the input and generated CSI using CrossEntropy as $L_4,L_5$. Additionally, we calculate all the loss functions focusing only on the [MASK] tokens again to ensure that the model prioritizes the recovery of the missing CSI.

\subsubsection{Recovering Phase}
The Recovering Phase is illustrated in Fig. \ref{Recovery Phase}. First, all the [PAD] tokens are replaced by the [MASK] token to make the data more similar to the form in pre-training. Then, the modified CSI is inputted into CSI-BERT.

Here, we propose two recovery methods named ``recover'' and ``replace''. The ``recover'' method is similar to traditional interpolation methods that only fill the lost tokens, while other tokens retain their original values. However, the Recoverer outputs a complete CSI sequence regardless of whether the token is [MASK], [PAD], or others. With the assistance of the Discriminator, the output sequence also maintains coherence. Therefore, we directly use the output sequence of the Recoverer as the recovered CSI. Since it is not identical to the actual CSI at each position, we refer to this method as ``replace''. The expression of ``recover'' and ``replace'' CSI is shown in Eq (\ref{recover}).

\begin{equation}
\begin{aligned}
& CSI^{replace}=\hat{C} , \\
& CSI^{recover}=(1-IsPad) \cdot C + IsPad  \cdot \hat{C} , \\
& IsPad=(C==[PAD]) , 
\label{recover}
\end{aligned}
\end{equation}
where $IsPad$ indicates whether the position of the original CSI $C$ corresponds to the [PAD] token.

\subsubsection{Fine-tuning Phase}
During the fine-tuning phase, we adopt the approach commonly used for pre-trained models, where we freeze the bottom layers of CSI-BERT and only train the top layers for specific classification tasks.

\section{Dataset and Numerical Analysis}
\subsection{Dataset}

In our study, we utilized the ESP32S3 microcontroller and a home Wi-Fi network to collect data at a sampling rate of 100Hz. The home Wi-Fi router served as the transmitter, while the ESP32S3, equipped with an external antenna, received CSI embedded in Ping Replies. This setup, shown in Fig. \ref{Workflow}, was cost-effective and operated at a carrier frequency of 2.4 GHz. Data collection took place in a cluttered conference room, with the transmitter and receiver positioned 1.5 meters apart. Eight volunteers, with an average age of 23.38 years, height of 1.75m, and BMI of 23.07, participated in the study. They performed various gestures such as moving left-right, forward-backward, up-down, circling clockwise, clapping, and waving. Each action was recorded for one minute, resulting in 60 samples per action. To construct our training dataset, we divided the CSI data within each second into individual samples. Our analysis revealed an average loss rate of 14.51\% in the collected CSI data. Across the 52 subcarriers, the maximum and minimum loss rates within each second were 70\% and 1\%, respectively.

\subsection{Experiment Setup}

\begin{table}
    \centering
        \begin{tabular}{|c|c|}
        \hline
        \textbf{Configuration} & \textbf{Our Setting} \\
        \hline 
        Input Length     & 100   \\
        \hline 
        Input Dimension     & 52   \\
        \hline
        Network Layers   & 4   \\
        \hline 
        Hidden Size   & 64    \\
        \hline 
        Inner Linear Size     & 128   \\
        \hline 
        Attn. Heads   & 4   \\
        \hline 
        Dropout Rate  & 0.1    \\
        \hline 
        Optimizer    & Adam \\
        \hline 
        Learning Rate    & 0.0005  \\
        \hline 
        Batch Size   & 64  \\
        \hline 
        Total Number of Parameters  & 2.11 M   \\
        \hline 
        \end{tabular}
\caption{Model Configurations: This table provides a detailed overview of our CSI-BERT in the following experiments. In this context, `M' represents `million' and `K' represents `kilo', which will remain consistent in the subsequent tables.}
\label{configuration}
\end{table}

We illustrate our model configurations in Table\ref{configuration}. We trained our model on an NVIDIA RTX 3090. During training, we observed that CSI-BERT occupied approximately 1300 MB of GPU memory.

\subsection{Experiment Result}
\begin{figure*}
\centering 
\subfloat[MSE]{\label{MSE}\includegraphics[width=0.18\textwidth]{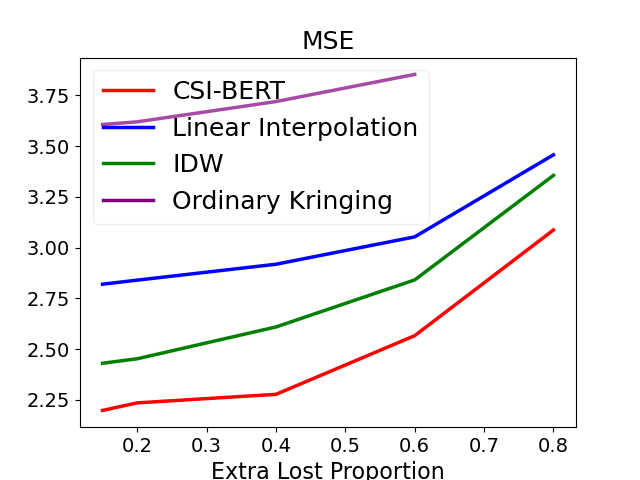}}
\subfloat[MAE]{\label{MAE}\includegraphics[width=0.18\textwidth]{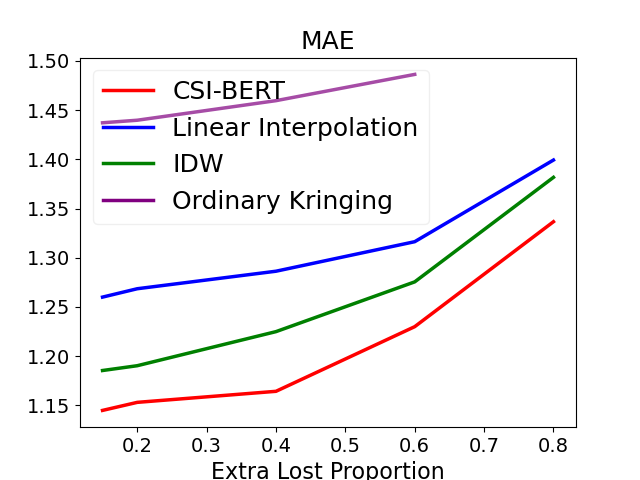}}
\subfloat[SMAPE]{\label{SMAPE}\includegraphics[width=0.18\textwidth]{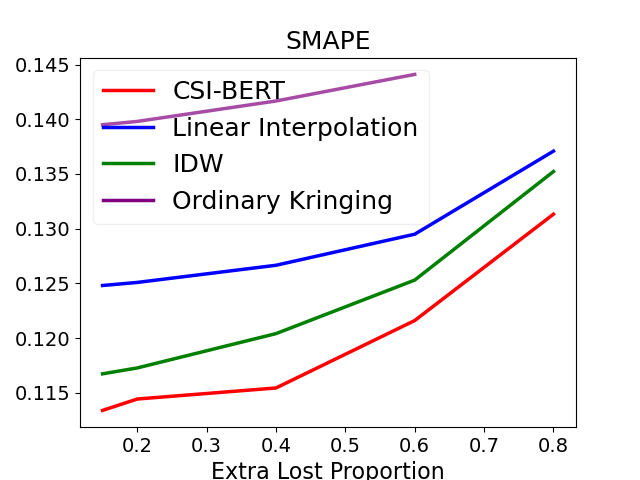}}
\subfloat[MAPE]{\label{MAPE}\includegraphics[width=0.18\textwidth]{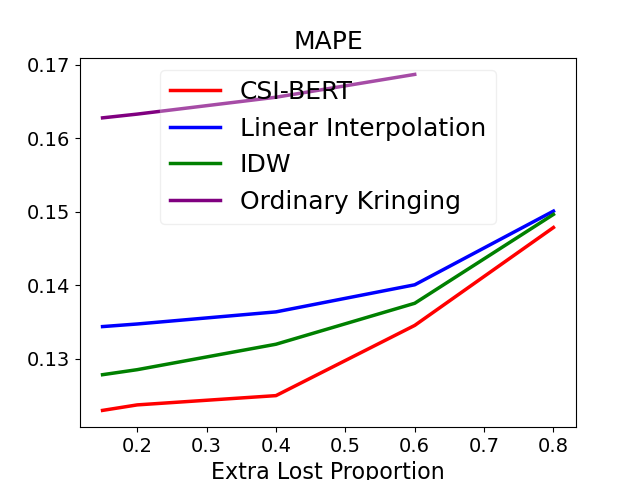}}
\subfloat[FSS]{\label{FSS}\includegraphics[width=0.18\textwidth]{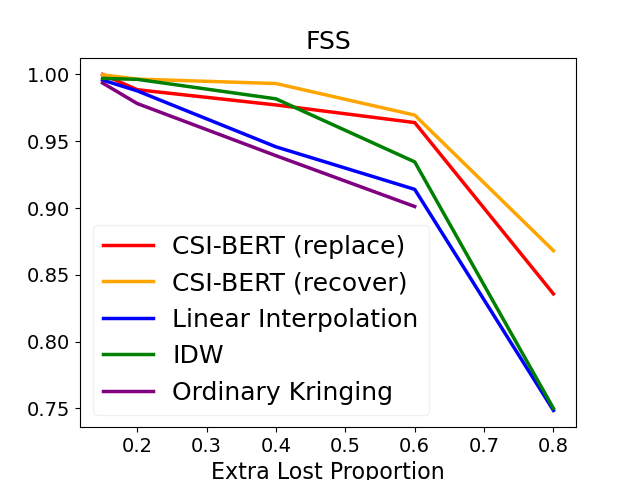}}
\caption{CSI Recovery Performance at Different Loss Rates: The x-axis represents the CSI loss rate (excluding the real lost CSI), and the y-axis represents the different metrics between the real CSI and CSI recovered by various methods. The line representing Ordinary Kriging is shorter than others, indicating its failure at higher loss rates.}
\label{delete}
\end{figure*}

In this section, we evaluate the recovery performance of CSI-BERT using two metrics. Firstly, we compare the Mean Squared Error (MSE), Mean Absolute Error (MAE), Symmetric Mean Absolute Percentage Error (SMAPE), and Mean Absolute Percentage Error (MAPE) between CSI-BERT and other interpolation methods including Linear Interpolation, Ordinary Kringing, Inverse Distance Weighted (IDW). To conduct this evaluation, we randomly delete 15\% of the non-[PAD] data from the testing set of CSI-BERT, and then use different methods to fill in these gaps. To ensure fairness, we calculate the aforementioned metrics only on the 15\% of the deleted CSI.
Besides, we also calculate the Fréchet Shape Similarity (FSS) \cite{FD} as shown in Eq (\ref{FSS_eq}) between whole original CSI and the recovered result.
\begin{equation}
\begin{aligned}
& \text{S}(C,\hat{C}) = \frac{\sum_{j=1}^d  S(C^{(j)},\hat{C}^{(j)})}{d} , \\
& \text{S}(C^{(j)},\hat{C}^{(j)})= \\ 
& \text{max}\{1-\frac{\underset{\alpha,\beta}{\text{inf}}\underset{t\in[0,1]}{\text{max}}D(C^{(j)}(\alpha(t)),\hat{C}^{(j)}(\beta(t)))}{\sqrt{\frac{1}{2}*||C^{(j)}||*||\hat{C}^{(j)}||}+\epsilon},0\} , 
\label{FSS_eq}
\end{aligned}
\end{equation}
where $d$ is the dimension of the flattened CSI $c_i$, as described in Eq (\ref{time embedding}). The functions $\alpha$ and $\beta$ are arbitrary continuous non-decreasing functions mapping from the interval $[0,1]$ to $[a,b]$, where $[a,b]$ represents the time range of the CSI. $D(\cdot,\cdot)$ denotes the Euclidean distance, $||C^{(j)}||$ and $||\hat{C}^{(j)}||$ represent the lengths of the curves $C^{(j)}$ and $\hat{C}^{(j)}$, respectively. What's more, we also compare the recovering time of different models. For fairness, we only use CPU during the comparison. The results are presented in Table \ref{error}. It is evident that CSI-BERT demonstrates the best performance across all metrics. Notably, we observe that the ``replace'' method of CSI-BERT achieves a higher FSS, compared to the ``recover'' method. This discrepancy may be attributed to the fact that the output of CSI-BERT exhibits inherent coherence, aided by the Discriminator. Additionally, to simulate a scenario with a higher loss rate, we manually deleted some CSI packages as shown in Fig. \ref{delete}. It is worth mentioning that CSI-BERT maintains its superior performance even with increased data loss rates. 

\begin{table*}
    \centering
        \begin{tabular}{|c|c|c|c|c|c|c|}
        \hline
        \textbf{Method} & \textbf{MSE $\downarrow$} & \textbf{MAE $\downarrow$} & \textbf{SMAPE $\downarrow$} & \textbf{MAPE $\downarrow$}  & \textbf{FSS $\uparrow$} & \textbf{Time Cost (min) $\downarrow$}\\
        \hline 
        CSI-BERT & \textbf{1.7326}  & \textbf{0.9413}  & \textbf{0.0902} &  \textbf{0.0945}  & \textbf{0.9999 (replace), 0.9979 (recover)} & \textbf{0.03}\\
        \hline 
        Linear Interpolation & 2.8294  & 1.2668  & 0.1248 & 0.1344 & 0.9841 &  0.64 \\
        \hline 
        Ordinary Kringing & 3.6067  & 1.4371 & 0.1627 & 0.1395 & 0.9936 &  45.15\\
        \hline 
        IDW & 2.4306  & 1.1854 &  0.1278 & 0.1167 & 0.9970 & 3.30\\
        \hline 
        \end{tabular}
\caption{Recovery Error: The bold value indicates the best result, which remains consistent across subsequent tables.}
\label{error}
\end{table*}

Then, we evaluate the performance of our recovered data and the interpolated data by inputting them into other models, including shallow Feed-Forward Network (FFN), shallow Convolutional Neural Network (CNN), RNN \cite{RNN}, Long Short-Term Memory (LSTM) \cite{LSTM}, and ResNet \cite{Resnet}, for two specific tasks: action classification and people identification. The results are presented in Table \ref{models}. It is evident that the data recovered by CSI-BERT achieves the highest improvement across most models. Furthermore, CSI-BERT itself demonstrates the best performance when the original data is used as input.

\begin{table*}
    \centering
        \begin{adjustbox}{width=\textwidth, height=0.08\textwidth}
        \begin{tabular}{|c||c|c|c|c|c|c||c|c|c|c|c|c|c|}
        \hline
        \textbf{Task} & \multicolumn{6}{c||}{\textbf{Action Classification}} & \multicolumn{6}{c|}{\textbf{People Identification}}\\
        \hline
        \multirow{2}{*}{\textbf{\diagbox{Data}{Model}}} & \textbf{FFN} & \textbf{CNN} & \textbf{RNN \cite{RNN}} & \textbf{LSTM \cite{LSTM}} & \textbf{ResNet \cite{Resnet}} & \textbf{CSI-BERT} & \textbf{FFN} & \textbf{CNN} & \textbf{RNN \cite{RNN}} & \textbf{LSTM \cite{LSTM}} & \textbf{ResNet \cite{Resnet}} & \textbf{CSI-BERT}\\
        &   337K & 23K & 33K & 133K & 11M &2M &   337K & 23K & 33K & 133K & 11M &2M\\
        \hline 
        Original Data & 66.93\% & 55.72\% & 39.56\% & 11.97\% & 70.31\% & \underline{76.91\%} & 71.34\% & 71.14\% & 66.39\% & 21.09\% & 83.76\% & \underline{93.94\%}\\
        \hline 
         CSI-BERT  recover & 74.23\% & 59.39\% & 48.96\% & 22.92\% & \textbf{\underline{92.57\%}} & 71.87\%  & \underline{97.13\%} & 80.60\% & 80.51\% & \textbf{35.18\%} & 94.30\% & 95.05\%\\
         \hline 
         CSI-BERT replace & \textbf{\underline{86.90\%}} & \textbf{61.51\%} & \textbf{58.80\%} & \textbf{52.36\%} & 84.52\% & \textbf{79.54\%} & \textbf{\underline{97.65\%}} & 79.18\% & \textbf{89.24\%} & 24.22\% & 97.39\% & 95.83\%\\
        \hline 
         Linear Interpolation & 72.91\% & 58.35\%  & 45.32\% & 49.09\% & \underline{80.75\%} & 74.55\% & 81.84\% & 70.88\% & 84.45\% & 26.83\% & 86.75\% & \textbf{\underline{97.92\%}}\\
        \hline 
        Ordinary Kringing & 65.62\% & 57.55\% & 53.64\% & 50.00\% & \underline{88.71\%} & 74.27\% & 94.76\% & \textbf{85.38\%} & 86.42\% & 21.61\% & \underline{97.32\%} & 95.83\%\\
        \hline 
        IDW & 40.17\% & 56.77\%  & 48.70\% & 46.88\% & \underline{80.32\%} & 67.22\% & 83.22\%  & 74.56\% & 88.54\% & 33.91\% & 94.27\% & \underline{95.20\%} \\
        \hline 
        \end{tabular}
        \end{adjustbox}
\caption{CSI Sensing Classification Performance: The number below each model name represents the number of parameters. The bold value indicates the best result in each column, and the underlined value indicates the best result in each row within each task.}
\label{models}
\end{table*}




\subsection{Ablation Study}
\begin{figure}
\centering 
\includegraphics[width=0.45\textwidth]{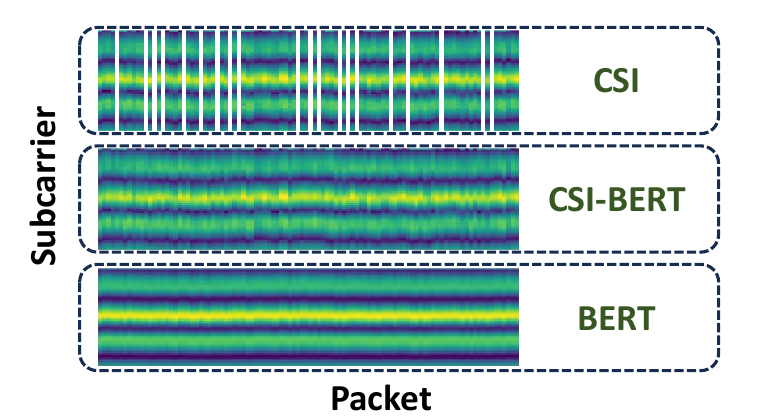}
\caption{Amplitude Spectrum of Original CSI and Output of BERT and CSI-BERT: The blank in the original CSI represents the lost CSI.}
\label{ablation fig}
\end{figure}

In this section, we conduct an ablation experiment to demonstrate the effectiveness of the modifications we made to BERT \cite{BERT}. We compare the recovery performance of the original BERT with our modified version, CSI-BERT, as shown in Fig. \ref{ablation fig}. We observe that the amplitude spectrum of the original BERT appears smooth across all subcarriers, indicating that it tends to map all tokens within each subcarrier to similar value. Although BERT can also achieve a relatively low MSE of 5.26, it fails to capture any valuable information.


\section{Conclusion}

In this paper, we present CSI-BERT, a novel model specifically designed for CSI recovery and sensing. CSI-BERT is the first model to be developed exclusively for CSI recovery. Our approach not only demonstrates the applicability of NLP models in the field of CSI but also leverages time information in CSI sensing, recovers multi-dimensional continuous data, and incorporates a novel pre-training method.

Through extensive experimentation, we have observed that CSI-BERT achieves outstanding performance in CSI recovery. The recovered data obtained with CSI-BERT not only enhances the performance of other CSI sensing models but also showcases superior performance in CSI sensing tasks. Besides, its fast recovering speed makes it suitable for real-time systems.


\section*{Acknowledgment}
This work is supported by Guangdong Major Project of  Basic and Applied Basic Research (No. 2023B0303000001), the Major Key Project of PengCheng Laboratory under grant PCL2023AS1-2, National Natural Science Foundation of China under Grant 62371313, Guangdong Basic and Applied Basic Research Foundation under Grant 2022A1515010109, the Internal Project Fund from Shenzhen Research Institute of Big Data under Grants J00120230001.

\bibliographystyle{ieeetr}
\bibliography{ref.bib}
\end{document}